\documentclass[10pt,twocolumn,letterpaper]{article}

\usepackage{iccv}
\usepackage[normalem]{ulem}

\usepackage{stfloats}
\usepackage{arydshln}
\usepackage{multirow}
\usepackage{booktabs}
\usepackage{float}

\definecolor{myPurple}{rgb}{0.4, .0, .8}
\definecolor{myGreen}{rgb}{0, 0.6, .3}
\definecolor{myRed}{rgb}{0.8, .2, .2}
\definecolor{myOrange}{rgb}{0.8, 0.45, 0.0}
\definecolor{myBlue}{rgb}{.0, .0, 1.0}
\definecolor{myBlue2}{rgb}{.0, 1.0, 1.0}
\definecolor{myBlack}{rgb}{.0, .0, 0.0}
\definecolor{darkmidnightblue}{rgb}{0.0, 0.2, 0.4}

\definecolor{MyGreen}{rgb}{0.02,0.5,0.02}

\newcommand{\method}{InstaScene\xspace}

\definecolor{iccvblue}{rgb}{0.21,0.49,0.74}
\usepackage[pagebackref,breaklinks,colorlinks,allcolors=iccvblue]{hyperref}

\def\confName{ICCV}
\def\confYear{2025}

\title{
InstaScene: Towards Complete 3D Instance Decomposition and \\ Reconstruction from Cluttered Scenes
\vspace{-1.0 em}
}

\author{
Zesong Yang$^{1,2}$\footnotemark[1] \qquad
Bangbang Yang$^{2}$ \qquad
Wenqi Dong$^{1,2}$ \qquad
Chenxuan Cao$^{1}$ \qquad
\\
Liyuan Cui$^{1}$ \qquad
Yuewen Ma$^{2}$ \qquad
Zhaopeng Cui$^{1}$ \qquad
Hujun Bao$^{1}$\footnotemark[2]
\vspace{0.2cm}
\\
{$^{1}$ State Key Lab of CAD \& CG, Zhejiang University\quad
 $^{2}$ ByteDance}
\\
\vspace{0.25 em}
\small{Project Page: \url{https://zju3dv.github.io/instascene/}}
}
\begin{document}

% adjust equation
\setlength{\abovedisplayskip}{0.2 em}
\setlength{\belowdisplayskip}{0.2 em}

\twocolumn[{
\renewcommand\twocolumn[1][]{#1}
\maketitle
    \vspace{-2.0 em}
    \centering
    \includegraphics[width=\textwidth]{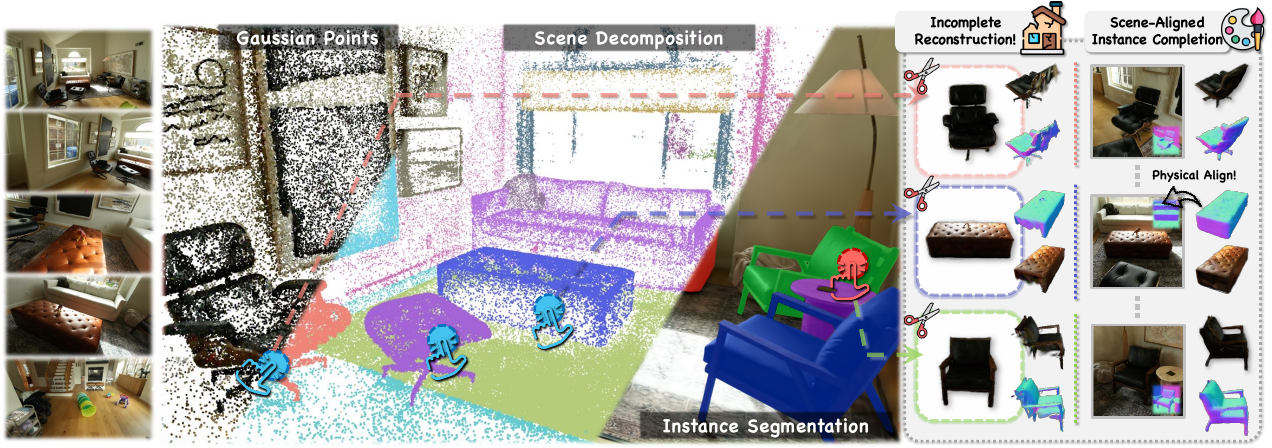}
    \captionof{figure}{
    \textbf{\method} allows users to pick up and decompose arbitrary instances from cluttered environments, while automatically reconstructing them into complete objects with intact geometry and appearance that align with the physical world.
    }
    \label{fig:teaser}
    \vspace{2.0 em}
}]

\maketitle

\renewcommand{\thefootnote}{\fnsymbol{footnote}}
\footnotetext[1]{\noindent Work done during an internship at PICO, ByteDance.}
\footnotetext[2]{\noindent Corresponding author.}

\begin{abstract}
Humans can naturally identify and mentally complete occluded objects in cluttered environments.
However, imparting similar cognitive ability to robotics remains challenging even with advanced reconstruction techniques, which models scenes as undifferentiated wholes and fails to recognize complete object from partial observations.
In this paper, we propose \textbf{\method}, a new paradigm towards holistic 3D perception of complex scenes with a primary goal: decomposing arbitrary instances while ensuring complete reconstruction.
To achieve precise decomposition, we develop a novel spatial contrastive learning by tracing rasterization of each instance across views, significantly enhancing semantic supervision in cluttered scenes.
To overcome incompleteness from limited observations, we introduce in-situ generation that harnesses valuable observations and geometric cues, effectively guiding 3D generative models to reconstruct complete instances that seamlessly align with the real world.
Experiments on scene decomposition and object completion across complex real-world and synthetic scenes demonstrate that our method achieves superior decomposition accuracy while producing geometrically faithful and visually intact objects.
\end{abstract}
\vspace{-1.5 em}
\section{Introduction}
\label{sec:intro}
\vspace{-0.5 em}

As humans, we possess an innate ability to understand cluttered 3D scenes and interact with diverse objects without deliberate observation.
Imagine walking into a crowded kitchen: we can immediately recognize each piece of furniture and every dish, and without a second thought you might reach out to pick up a specific utensil from a busy countertop.
Over the years, numerous efforts have been made in computer vision and robotics to achieve similar skills~\cite{kroemer2021review}, developing series of capabilities like efficient reconstruction and rendering~\cite{kerbl20233d}, 3D scene understanding~\cite{peng2023openscene}, and 3D object generation~\cite{poole2022dreamfusion,hong2023lrm,zhang2024clay}.
However, the gap for downstream scene editing and simulation applications~\cite{ji2024graspsplats,shorinwa2024splat, zhou2024drivinggaussian,yan2024street} still remains.

Several key limitations in existing approaches hinder this goal.
First, most generic 3D reconstruction methods~\cite{mildenhall2021nerf,wang2021neus,schonberger2016structure,kerbl20233d} usually treat the scene as a whole model, which hinders delicate instance-level tasks such as manipulation and rearrangement;
Second, open-set scene understanding methods~\cite{peng2023openscene,takmaz2023openmask3d} enables object-level query and segmentation but fails to produce complete instances from cluttered environments;
Third, category-specific generative approaches can predict complete 3D shapes from partial observations, but they often struggle to generalize to diverse objects in cluttered scenes, or ensure physical and visual alignment with the real world \cite{liu2024lasa}, \ie, matching the object's actual size, shape, and texture, leading to physical and visual inconsistencies.

To enable robotic systems with human-like perception for understanding and interacting with surrounding environment, a new paradigm is required to seamlessly integrate 3D understanding, reconstruction, and generation into a unified system with two essential capabilities:
\textbf{1) Precise instance decomposition}: the system should precisely segment and isolate arbitrary object instances in a complex scene, irrespective of their categories.
\textbf{2) Complete instance-aware scene reconstruction}: for every object that is extracted, the system should reconstruct the object's complete geometry and appearance while adhering to the physical world, even if parts of it were never directly observed due to occlusion or limited viewpoints. 

\begin{figure}[t] 
    \centering
    \scriptsize
    \includegraphics[width=0.475 \textwidth]{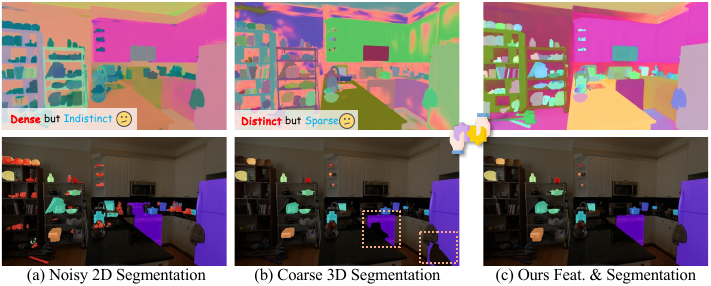}
    \caption{
    \textbf{Motivation of spatial contrastive learning with mutual guidance.} 
    In complex scenes, naively supervising feature field with noisy 2D segmentation masks results in indistinct features (see (a)), and only using 3D masks from spatial trackers would result in sparse Gaussian points (see (b)).
    Observed that the former provides dense features while the latter offers a robust reference, we leverage the interplay to mutually guide each other and achieve a dense and distinct feature field (see (c)).
    }
    \label{fig:motivate}
\end{figure}

In this work, we propose \textbf{\method}, a framework that realizes the above vision by integrating instance segmentation and complete reconstruction in a unified pipeline.
At a high level, our method takes a captured cluttered scene (\ie, in the form of Gaussian Splatting~\cite{kerbl20233d}), and decomposes it into complete instances that possess faithful geometry and appearance.
By addressing segmentation and reconstruction together, \textbf{\method} ensures that the object in the scene can be individually obtained as a complete 3D model \textbf{aligned} with the scene's context.
To achieve this goal, we first develop a novel spatial contrastive learning scheme guided by traced Gaussian clustering for fine-grained instance decomposition.
Specifically, we construct spatial trackers that identify and cluster the Gaussian points primarily contributing to each mask's rasterization, enabling consistent instance identification across multiple views. 
These trackers provide reliable supervision for the instance-level feature field through spatial contrastive learning, yielding highly distinguishable features as shown in Fig.~\ref{fig:motivate} (c).
This enables precise scene decomposition even in challenging environments (\eg, closely spaced bottles in Fig.~\ref{fig:exp:zipnerf_compare}).

To recover complete objects that align with real-world scenes, a primary challenge is that clean object views required by typical 3D generative models~\cite{poole2022dreamfusion,hong2023lrm,zhang2024clay} are rarely available in complex scenes, where objects are often partially occluded or captured from suboptimal views.
To tackle this challenge, we propose in-situ generation, which harnesses a 3D generative model to leverage valuable information from known observations and partially reconstructed geometry as omni-conditions.
Specifically, for each instance, we employ occlusion-aware viewpoint selection to render optimal views that serve as alternated conditions for the diffusion process.
We further incorporate geometry hints to complement known latent features via feature warping, enhancing multi-view consistency between observations and generations.
The generated views, together with source observations, are then used to fine-tune the object's Gaussian model, achieving complete reconstruction that can be directly placed back into the scene (referred to as "in-situ" or "in place").
Through these comprehensive designs, our method effectively emulates the human cognitive ability to identify and mentally complete partially observed objects in complex environments, thereby facilitating downstream tasks like robot-object interaction.

\begin{figure*}[t] 
    \centering
    \scriptsize
    \vspace{-0.75 em}
    \includegraphics[width=0.975 \textwidth]{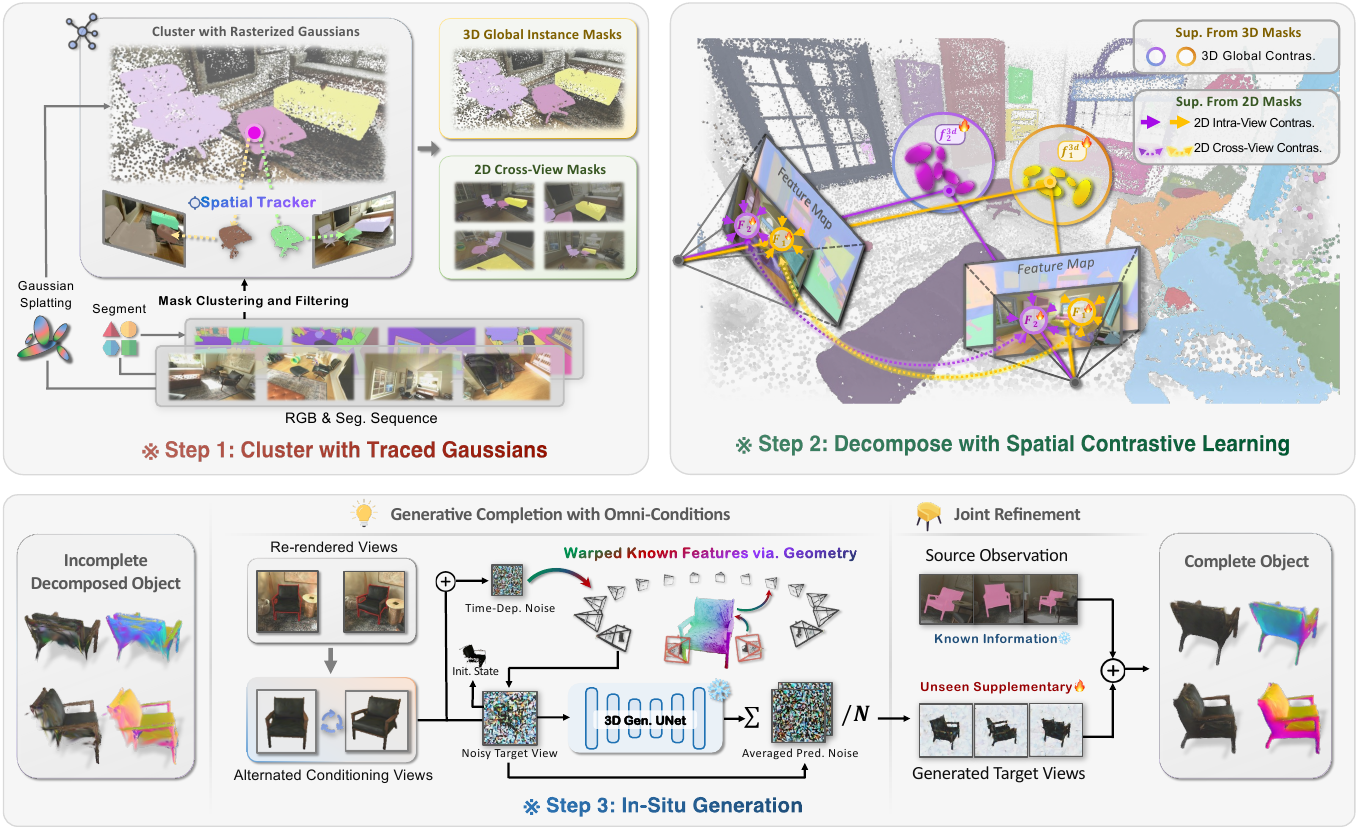}
    \caption{\textbf{System Overview.}
    Given a reconstructed Gaussian Splatting scene, our method first clusters and filters 2D segmentation masks by tracing the rasterization of Gaussian Splatting, which yields 2D and 3D instance masks.
    Then, we use spatial contrastive learning with mask supervision to train a feature field that achieves high-quality scene decomposition.
    Finally, for each decomposed incomplete object, we conduct an in-situ generation pipeline that takes all known observations and geometric cues as omni-conditions to the 3D generative model to obtain supplemented views, which will be jointly fine-tuned with source training views to obtain a complete object.
    }
    \vspace{-0.25 em}
    \label{fig:pipeline}
\end{figure*}

Our main contributions can be summarized as follows:
\begin{itemize}
    \item We introduce \textbf{\method}, a novel framework, which decomposes arbitrary objects from complex scenes while ensuring complete geometry and appearance reconstruction of each instance with limited observation.
    \item We propose a novel spatial contrastive learning scheme based on a traced Gaussian clustering technique, achieving accurate scene decomposition.
    \item We design a new in-situ generation pipeline, which aggregates all available observations and geometric hints as omni-conditions to control the 3D generative prior, yielding complete instance modeling that seamlessly aligns with the physical world.    
    \item  Extensive validations on various complex datasets demonstrate the superior performance of our method in fine-grained scene decomposition and faithful complete object reconstruction, with promising applications for downstream tasks like scene editing and manipulation. 
\end{itemize}

% \vspace{-0.75 em}
\section{Related Work}
\label{sec:relatedwork}
\vspace{-0.5 em}
\noindent\textbf{Instance-aware Scene Reconstruction.}
Generic scene reconstruction approaches~\cite{mildenhall2021nerf,kerbl20233d,schoenberger2016mvs} usually model the entire scene as a single representation.
For further scene understanding tasks, recent works tend to conduct reconstruction at instance-level granularity~\cite{yang2021objectnerf, li2023rico, wu2022object, liu2024lasa,wu2023objectsdf++}.
However, existing methods either require laborious manual mask annotation~\cite{yang2021objectnerf,wu2022object,wu2023objectsdf++,li2023rico} 
for instance decomposition 
and usually encode instances through multiple MLPs, which struggle with the scene containing numerous objects due to the limited scalability.
Another line of work learns a generative shape prior from predefined CAD models or large-scale synthetic datasets~\cite{liu2024lasa,fu20213d,avetisyan2019scan2cad,chang2015shapenet}.
Nevertheless, due to the limitation of 3D data collection, these methods often face difficulties in generalizing to complex real-world scenes with numerous categories and diverse shapes, which leads to domain gaps and shape misalignment between reconstructed results and real-world objects.
Besides, most of these works only focus on geometric modeling while ignoring realistic appearance rendering, which restricts the applications in vision-based simulation and scene editing.

% \vspace{-0.25 em}
\noindent\textbf{3D Segmentation with Scene Reconstruction.}
Recent advances in large-scale vision, language, and segmentation models~\cite{radford2021learning,caron2021emerging,oquab2023dinov2,kirillov2023segment} have enabled encoding rich visual features into volumetric fields or 3D Gaussians,
facilitating versatile 3D instance segmentation over existing scene reconstruction.
These works either reduce the dimension of pre-trained features with quantization and distillation~\cite{qiu2024feature,zhou2024feature, shi2024language, qin2024langsplat}, or conduct video tracking for cross-view masks association~\cite{ye2025gaussian}.
However, VLM-based semantic features struggle with complex, cluttered scenes containing duplicate objects (Fig.~\ref{fig:exp:zipnerf_compare}), while video tracking becomes unreliable under heavy occlusion, which inevitably messes up the feature learning. 
Some approaches \cite{kim2024garfield,choi2025click,wu2024opengaussian,ying2024omniseg3d,liu2024part123} adopt contrastive learning scheme, which enables object extraction from well-observed data but still struggles under cluttered scenes due to inconsistent 2D priors (see Fig.~\ref{fig:motivate} in the experiments).
Besides, none of these works consider the completeness of the extracted instance, which limits the application in downstream scene editing or simulation tasks.

% \vspace{-0.5 em}
\noindent\textbf{3D Inpainting and Amodal Reconstruction.}
To obtain complete instances under insufficient observation or severe occlusion, one line of works is to apply 3D inpainting.
However, most related works focus on scene-level inpainting using 2D inpainting tools~\cite{mirzaei2023spin,suvorov2022resolution,lugmayr2022repaint,liu2024infusion}, which excels at removing objects from a clean background but cannot recover occluded parts of complex objects.
For object-level inpainting tasks, \cite{hu20242, weber2024nerfiller} use 2D diffusion-based inpainting tools but require user-provided mask sketches and prompts, and the result is not stable.
Apart from that, one promising approach is to utilize recent image-to-3D models~\cite{liu2023zero,shi2023zero123++,liu2023syncdreamer,hu2024mvd, dong2024coin3d} for generative reconstruction, but generic generation pipelines cannot ensure alignment with real-world scenes or properly handle incomplete/occluded observation (see Sec.~\ref{exp:complete}), limiting the applicability for precise and faithful reconstruction.
A similar concurrent work, DP-Recon~\cite{ni2025dprecon}, employs generative priors to improve sparse and occluded regions—geometry completion then texture refinement, 
while our method jointly completes both in a single step with realistic rendering.

% \vspace{-0.75 em}
\section{Methods}
\label{sec:methods}
\vspace{-0.5 em} 
We present a novel framework for instance-aware reconstruction in complex scenes, jointly addressing segmentation and instance completion in a unified pipeline. 
Unlike prior approaches that treat segmentation and generation as disjoint tasks, our segmentation stage provides essential spatial priors (e.g., geometry, object-centric views, masks) that directly guide the subsequent generative module.  Our key components include a spatial contrastive learning for fine-grained scene decomposition (Sec.~\ref{sec:scene_decom}), and a new in-situ generation for complete instance modeling (Sec.~\ref{sec:insitu_gen}).

\begin{figure}[!t] 
    \centering
    \scriptsize
    \includegraphics[width=0.45 \textwidth]{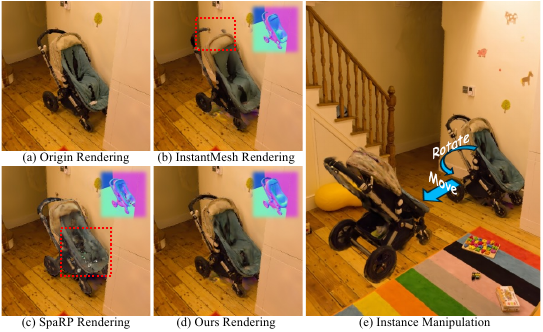}
    \caption{\textbf{In-Situ Generation vs. Generic Image-to-3D.}
    We perform different instance-level complete reconstruction approaches on the baby carriage and put it back on the scene.
    Generic image-to-3D methods~\cite{xu2024instantmesh,xu2025sparp} struggle to maintain consistent reconstruction with the original scene (a) and suffer from misalignment (such as broken handles or seats in (b) and (c)).
    Our in-situ generation ensures the appearance and geometric consistency to the original scene while faithfully completing the unseen regions (see (d)).
    We also show the application of scene manipulation on the decomposed complete instance in (e).}
    \vspace{-0.25 em} 
    \label{fig:insitu_show}
\end{figure}

% \vspace{-1.25 em}
\subsection{Preliminary}
\vspace{-0.25 em}
Given posed RGB image sequences as input, we first reconstruct the scene with point-based rendering method 2D Gaussian Splatting~\cite{huang20242d}. Building upon 3DGS~\cite{kerbl20233d}, 2DGS collapses the 3D ellipsoid volumes into a set of 2D oriented planar Gaussian disks, enhancing view-consistent rendering while improving the geometric quality.
A 2D Gaussian is parameterized as a local tangent plane, which is defined as:
\begin{align}
P(u,v) &= \mathbf{p}_k + \mathbf{s}_u \mathbf{t}_u u + \mathbf{s}_v \mathbf{t}_v v ,
\label{eq:plane-to-world}
\end{align}
where the center $\mathbf{p}_k$, scaling $(\mathbf{s}_u,\mathbf{s}_v)$, the rotation $(\mathbf{t}_u,\mathbf{t}_v)$, opacity $\alpha$ and view-dependent appearance $\mathbf{c}$ with spherical harmonics are learnable parameters. 
For a point $\mathbf{u}=(u,v)$ in $uv$ space, its 2D Gaussian value can be evaluated as:
\begin{equation}
\vspace{-0.25 em}
\mathcal{G}(\mathbf{u}) = \exp\left(-\frac{u^2+v^2}{2}\right).
\label{eq:gaussian-2d}
\end{equation}
Through rasterization, Gaussians passed through by the rays $\mathbf{x}$ emitted from the $uv$ space are composed into pixel appearance with depth-sorted alpha blending, as:
\begin{equation}
\vspace{-0.25 em}
\label{eq:2dgs-rasterizer}
\begin{split}
\mathbf{c}(\mathbf{x}) = \sum_{i=1} \mathbf{c}_i \alpha_i \mathcal{G}_i(\mathbf{u}(\mathbf{x})) \prod_{j=1}^{i-1} (1 - \alpha_j \mathcal{G}_j(\mathbf{u}(\mathbf{x}))).
\end{split}
\vspace{-0.25 em}
\end{equation}

We augment each Gaussian with a $D$-dimensional randomly initialized embedding ${\mathbf{f}}_{i}^{3d} \in {\mathbb{R}}^{D}$ as the instance-aware feature.
With the rasterizer akin to Eq.~\ref{eq:2dgs-rasterizer}, we obtain pixel-level features $\mathbf{f}$, as:
\vspace{-0.5 em}
\begin{equation}
\label{eq:2dgs-feat-rasterizer}
\mathbf{f}(\mathbf{x}) = \sum_{i=1} \mathbf{f}_i^{3d} \alpha_i \mathcal{G}_i(\mathbf{u}(\mathbf{x})) \prod_{j=1}^{i-1} (1 - \alpha_j \mathcal{G}_j(\mathbf{u}(\mathbf{x}))).
\end{equation}

In our setup, we set $D=16$ and freeze other attributes.

\vspace{-0.25 em}
\subsection{Scene Decomposition with Spatial Contrastive Learning}
\label{sec:scene_decom}
\vspace{-0.25 em}
\noindent\textbf{Mask Clustering with Spatial Gaussian Tracker.}
We first employ the off-the-shelf mask predictor EntitySeg~\cite{qi2022high} to generate class-agnostic instance-level 2D segmentation masks. 
To obtain 3D segmentation of the scene, existing methods~\cite{kim2024garfield,choi2025click,wu2024opengaussian,ying2024omniseg3d,liu2024part123} generally lift 2D segmentation priors to a unified 3D space.
However, 2D segmentation priors are usually noisy (\eg, cross-view inconsistency, under-segmentation), which introduces ambiguity and suboptimal supervision in feature field learning (see Fig.~\ref{fig:motivate}).
To extract robust supervision from 2D masks, 
inspired by previous works that exploit the re-projected spatial consistency of 2D segmentations to achieve robust 3D segmentation \cite{yang2023sam3d,yan2024maskclustering,nguyen2024open3dis,yin2024sai3d,lu2023ovir}, we apply a trustful mask clustering strategy with traced Gaussian points,
which back-projects the segmentation masks into 3D space and utilizes the spatial relationships among the corresponding Gaussian points to achieve cross-view mask matching, which yields a robust global semantic prior for scene decomposition.

Due to the inherently noisy distribution of reconstructed 3DGS point clouds, naive depth projection discards potentially valid points associated with 2D segmentations.
Given the frame $I_i$ along with its semantic map $M_i$, we render $\bar{I}_i$ with its corresponding pose and define the Gaussians that contribute to the rasterization of each pixel within the $j$-th mask $m_{i,j}$, with transmittance exceeding 0.5, as individual spatial tracker $P_{i,j}$.
To determine whether two masks belong to the same instance, we utilize the view consensus rate as proposed in \cite{yan2024maskclustering}. Specifically, a spatial tracker $P_{i,j}$ is considered visible at frame $I_{i'}$ if its 30\% points contribute to the rasterization of $I_{i'}$, and $P_{i,j}$ is contained within frame $I_{i''}$ if 80\% of its points appear within a tracker $P_{i'',j''}$.
Given any two mask trackers $P_{i,j}$ and $P_{k,l}$, the view consensus rate $\mathcal{C}$ is defined as:
\begin{equation}
    \label{eq:view_consensus_rate}
    \mathcal{C}(P_{i,j},P_{k,l})=\frac{N_{contain}(P_{i,j},P_{k,l})}{N_{vis}(P_{i,j},P_{k,l})},
\end{equation}
where $N_{vis}$ and ${N_{contain}}$ represent the number of frames where both trackers are visible and contained respectively.
If $\mathcal{C}$ exceeds 0.9, the corresponding masks $m_{i,j}, m_{k,l}$, can be considered to belong to the same instance.

Additionally, we categorize $m_{i,j}$ as under-segmentation if its associated spatial tracker $P_{i,j}$ simultaneously intersects with multiple trackers $\{P_{k,j'}\}$ from the same frame $I_k$ and is consistently present in its visible frames. Then we discard $m_{i,j}$. Thus, we achieve cross-view mask clustering and filter noisy segmentation by tracing rasterization of Gaussians. We denote the clustered cross-view segmentations for instance $\mathcal{I}_n$ as $\mathcal{M}^{2d}_n = \{m_{i,j}\}$.
In addition to utilizing these multi-view consistent semantic priors, we merge Gaussian tracker points belonging to the same instance and employ DBSCAN\cite{ester1996density} to filter out floaters. The fused Gaussians $\mathcal{P}_n$ serve as a robust 3D global instance mask $\mathcal{M}^{3d}_n$ for instance $\mathcal{I}_n$. 

\noindent\textbf{Spatial Contrastive Learning.}
Although DBSCAN is applied to remove floaters, it inadvertently discards semantically meaningful Gaussian points misclassified as noise (see Fig.~\ref{fig:motivate}.b). To address this, 
we propose a spatial contrastive learning framework that jointly leverages intra-view and cross-view consistent 2D segmentation masks, along with robust 3D global instance masks, to effectively guide the learning of a distinctive and semantically coherent feature field.
While the 3D masks are sufficiently distinguishable to guide the learning of 2D features with corresponding labels at a global level, the 2D segmentation masks supplement the Gaussians that are filtered out as floaters in 3D masks as shown in Fig.~\ref{fig:motivate}. This interplay allows the 2D local mask supervision and the 3D global mask supervision to mutually guide each other in learning a distinctive feature field.

\begin{figure*}[!t]
    \centering
    \scriptsize
    \vspace{-1.5 em}
    \includegraphics[width=\textwidth]{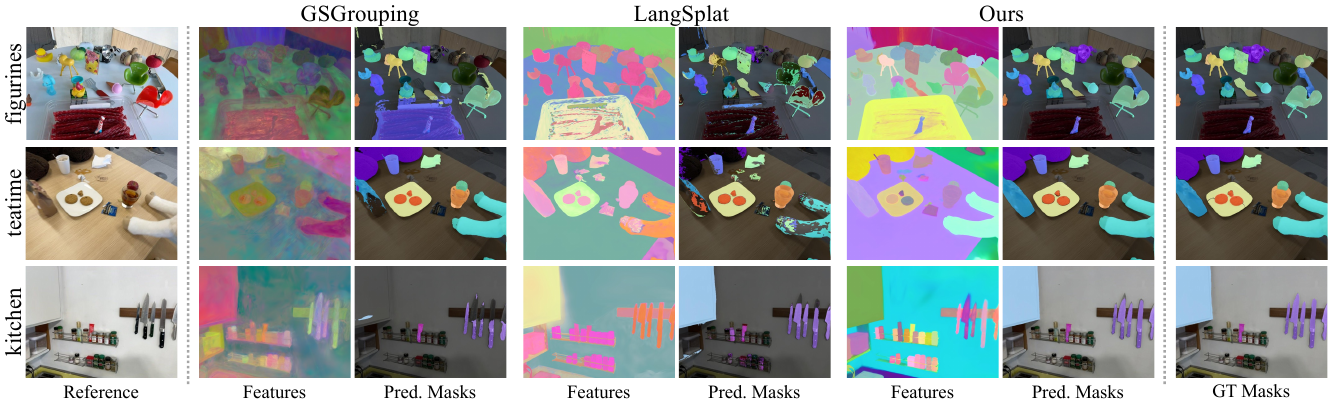}
    \caption{
    \textbf{Comparison on LERF-Mask Dataset.} 
    We compare the segmentation results and visualized features (with PCA) of GSGrouping~\cite{ye2025gaussian}, LangSplat~\cite{qin2024langsplat} and ours.
    Our method learns highly distinguishable features and achieves the most precise instance segmentation.
    }
    \vspace{-0.5 em}
    \label{fig:exp:lerf_compare}
\end{figure*}

\begin{table}[!t]
\resizebox{1.0\linewidth}{!}{
\tabcolsep 10pt
\huge
\begin{tabular}{ccccc}
\toprule
Methods     & Figurines & Teatime & Kitchen & Average \\
\midrule
Langsplat~\cite{qin2024langsplat} & 58.1      & 73.0    & 50.7   & 60.6    \\
GSGrouping~\cite{ye2025gaussian}   & 59.0      & 72.3    & 43.1    & 58.1  \\
Ours      & \textbf{85.7}      & \textbf{93.7}    & \textbf{77.3}    & \textbf{85.6}  \\
\bottomrule
\end{tabular}
}
\vspace{0.15em}
\caption{
    \textbf{Quantitative comparison on LERF-Mask Dataset.} We report mIoU~(\%) metric across three scenes. Our method demonstrates a substantial improvement over baselines in instance-level segmentation, especially on complex scenes.
}
\label{tab:compare_lerf}
% \vspace{-0.25 em}
\end{table}

Given features with instance labels $\mathcal{F}=\{f_{i}^j\}$, we use contrastive learning to maximize the similarity for features with the same semantic labels $\{f_i\}$ while distinguishing features from different labels with the following loss as:
\begin{equation}
\hspace{-0.6 em} 
\mathcal{L}_{CF}(\mathcal{F})=-\frac{1}{N}\sum_{i=1}^{N}\sum_{j=1}^{|\{f_i\}|}\log{\frac{\exp(f_{i}^{j}\cdot \bar{f}_i/ \phi_i)}{\sum_{k=1}^{N}\exp(f_{i}^{j}\cdot \bar{f}_k/ \phi_k)}},
\label{eq:cc}
\end{equation}
where $N$ is the number of instances involved in $\mathcal{F}$, $\bar{f}_i$ is the mean value for $f_{i}$, and $\phi_i$ is the instance temperature. 
We apply $\mathcal{L}_{CF}$ to features with corresponding labels sampled from the single view $(\mathbf{F}_i$, $M_i)$, adjacent views $\bar{\mathbf{F}}_i=\{(\mathbf{F}_j, \mathcal{M}_j^{2d}) \mid j \in [i-k, i+k]\}$, and 3D Gaussian points $(\mathbf{f}^{3d}_i,\mathcal{M}^{3d}_i)$, 
and train the feature field with loss:
\vspace{0.125 em}
\begin{equation} 
\mathcal{L}_{\mathcal{F}}=
\lambda_1 \mathcal{L}_{CF}(\mathbf{F}_i) + 
\lambda_2 \mathcal{L}_{CF}(\bar{\mathbf{F}}_i) + 
\lambda_3 \mathcal{L}_{CF}(\mathbf{f}^{3d}_i),
\label{eq:cf}
\vspace{0.125 em}
\end{equation}
where $\mathbf{f}^{3d}_i$ represents the features of visible Gaussians for frame $I_i$, and $\mathbf{F}_i$ is the rendered feature with Eq.~\ref{eq:2dgs-feat-rasterizer}. 
Upon completing the training, we perform instance segmentation by calculating cosine similarity between the average of coarse 3D instance features \(\hat{f}_i^{3d}\) and each Gaussian feature, applying $\tau_{seg}=0.9$ as the segmentation threshold.
Please refer to supp. material for specific implementation details.

\vspace{-0.5 em}
\subsection{In-Situ Generation}
\label{sec:insitu_gen}
\vspace{-0.25 em}
We aim for the reconstruction to be not only complete, even with occlusions and partial observations, but also authentic, featuring realistic appearance and close shapes that align with the real-world (as demonstrated in Fig.~\ref{fig:insitu_show}).
To fulfill all these demands, we propose a novel in-situ generation pipeline, which tames the generic 3D generative model with all known information to achieve realistic and complete instance reconstruction.

\begin{table}[t]
\resizebox{1.0\linewidth}{!}{
\tabcolsep 10pt
\begin{tabular}{lcccc}
\toprule
Model        & Figurines & Teatime & Kitchen & Average \\
\midrule
with $m_{noisy}^{2d}$    & 80.3     & 90.1   & 71.2    & 80.5   \\
\hdashline
with $M^{3d}$     & 81.5      & 88.5   & 67.0    & 79.0  \\
+ with $m_{filter}^{2d}$ & 83.9      & 91.4    & 75.4    & 83.6  \\
+ with $m_{cv}^{2d}$     & \textbf{85.7}      & \textbf{93.7}    & \textbf{77.3}    & \textbf{85.6}  \\
\bottomrule
\end{tabular}
}
\vspace{0.125 em}
\caption{\textbf{Ablation Study of spatial contrastive learning.} We gradually add 2D supervision into the spatial contrastive learning. $m_{filter}^{2d}$ represents contrastive learning on the filtered 2D intra-view masks, and $m_{cv}^{2d}$ denotes contrastive learning on the 2D cross-view masks.
}
% \vspace{-1.0 em}
\label{tab:ablation_seg}
\end{table}

\noindent\textbf{3D Generation with Omni-Conditions.}
We first formulate our diffusion process with omni-conditions.
As the generative models are already capable of inferring 3D geometric distributions from a single image, given all known information of the partial reconstructed instance $\mathcal{I}$ (\eg, views $\{y^{k}\}$ and depths $\{d^{k}\}$ rendered in visible viewpoints $\{\pi^{k}\}$), we aim to control the 3D diffusion model~\cite{hu2024mvd} with these available omni-conditions to predict the unseen regions $p(\{x^{n}\} | \{y^{k},d^{k},\hat{\pi}_n^{k}\})$ with precise alignment to real-world identity,
where $\{x^n\}$ is the unseen views in $\{\pi^{n}\}$, and $\{\hat{\pi}_n^{k}\}$ is the relative pose between input and target viewpoint.
To complement the generation model with known visual observations, we first design an alternated view conditioning strategy, which sequentially feeds optimal views re-rendered with instance's 2DGS as alternated conditions through the denoising process, as shown in the Step-3 of Fig.~\ref{fig:pipeline}.
For each timestep, we average the noise predictions across each target view $\{{\epsilon_\theta}_n^{k}\}$, as:
\begin{equation}
    \begin{split}
    x_{t-1}^n = \frac{1}{\sqrt{\alpha _t}}x_t^n-\frac{\beta_t}{\sqrt{1-\bar{\alpha} _t}}\bar{\epsilon}_\theta^n, \\
    \bar{\epsilon}_\theta^n  = \frac{1}{N_k}\textstyle\sum_{k=1}^{N_k}\epsilon_\theta^n(x_t^n,y^k,\hat{\pi}_n^k),
    \end{split}
\end{equation}
where $\epsilon_\theta$ is the noise predictor and $\bar{\epsilon}_\theta^n$ represents the averaged predicted noise for target view $x^n$. 

\noindent\textbf{Complement Known Features via Geometry Cues.}
Even with complemented views as conditions, the predicted views from the generative model may still deviate from the real observation due to the domain gap between training data and the real-world scene (see Fig.~\ref{fig:exp:gen_ab}).
To mitigate this, we leverage geometric cues from existing Gaussian scenes and design a geometry-aware feature warping strategy, which further enforces consistency from the known observation to the generative prediction.
During each diffusion iteration, time-dependent noise is added to the input views' latent features, which are then projected onto visible pixels of target views using rendering depths.
Since 2DGS yields a fused mesh, we use the surface normal of the viewing-ray-mesh intersection to discard projections indicating back-facing surface. 
This enforces consistency in visible regions, while invisible parts are initialized as random noise and are gradually denoised through guidance from both alternated condition views and the warped latent features from visible pixels.
See supp. for more details.

\begin{figure*}[!t]
    \centering
    \scriptsize
    \vspace{-0.5 em}
    \includegraphics[width=0.975\textwidth]{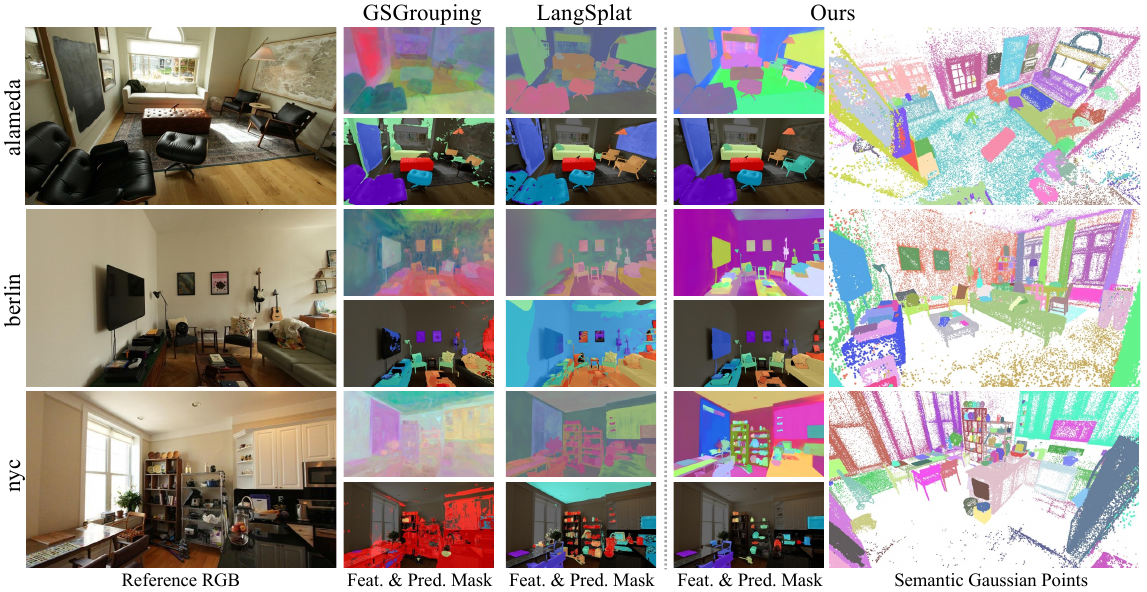}
    \caption{\textbf{Comparison on ZipNeRF Dataset.} The baselines yield highly noisy segmentation results in such complex scenes, while our method achieves fine-grained instance-level segmentation.
    We also show the segmented Gaussian points of our method.
    }
    \vspace{-0.75 em}
    \label{fig:exp:zipnerf_compare}
\end{figure*}

% \vspace{-0.25 em}
\noindent\textbf{Occlusion-Aware Viewpoint Selection and Joint Refinement.}
Unlike the typical image-to-3D pipeline that takes a standard elevated view as input, occlusions in cluttered scenes make the optimal viewpoints for each instance not straightforwardly available.
Hence, we set up the following principles that select the appropriate viewpoints as ideal input conditions for each instance. 
First, 
inspired by the target view setup in \cite{liu2023syncdreamer}, 
we select 16 viewpoints centered around the segmented object.
Among these viewpoints, those with the least scene occlusion between the viewpoint and the object are selected as the ideal input conditions, while the remaining viewpoints, subject to occlusion, are considered unseen and require supplementation from the 3D generative prior as explained above. We filter out the background in the selected rendering views with the 2D instance mask extracted from the rendered features.
Finally, we jointly refine the instance's 2DGS with both the source observations and the generated views.
The source observation ensures consistency for the visible parts, while the generated views are responsible for supplementing the unseen parts. 
Thus, we achieve a complete instance reconstruction guided by the known information.
See more details in supp. material.

\vspace{-2.0 em}
\section{Experiments}
\vspace{-0.5 em}
\label{sec:experiment}
We compare \method for instance-aware scene reconstruction in two folds. The first section focuses on fine-grained instance decomposition from scene reconstruction. The second section analyzes the in-situ generation for instance-level complete reconstruction, which measures the quality and accuracy of recovering decomposed objects.

\vspace{-0.25 em}
\subsection{Fine-Grained Instance Decomposition}
\vspace{-0.25 em}

\noindent\textbf{Datasets.} 
To demonstrate our performance in 3D instance segmentation, 
we conduct a quantitative comparison using LERF-Mask Dataset~\cite{kerr2023lerf}. We select three instance-rich scenes: figurines, waldo-kitchen, and teatime, and manually re-annotate them with instance-level ground truth masks.
Following \cite{choi2025click}, we extract the instance features from a given reference view, generate the corresponding instance masks in the target view using cosine similarity, and calculate mIoU between the extracted masks and groundtruth masks as the evaluation metric. 
To further demonstrate the robustness of our method in complex environments, we present qualitative comparison results with three scenes selected from the ZipNeRF Dataset~\cite{barron2023zip}, which feature a variety of objects.
We also conduct a comparison on 3D-OVS Dataset~\cite{liu2023weakly} in supp. material.

\begin{figure*}[!t] 
    \centering
    \scriptsize
    % \vspace{-5.5 em}
    \includegraphics[width=\textwidth]{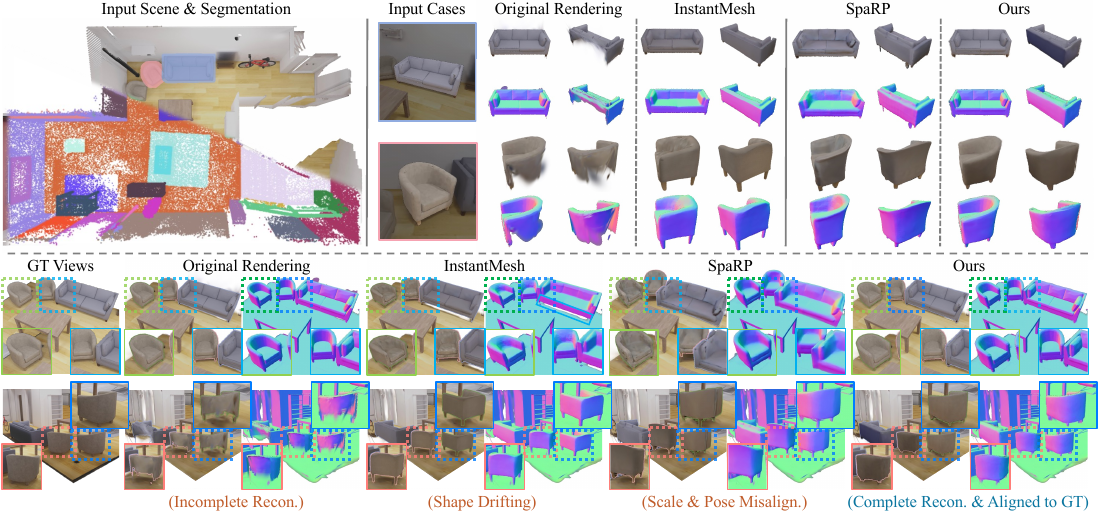}
    \caption{\textbf{Comparisons on Replica-CAD Dataset~\cite{szot2021habitat}.} 
    Instances extracted from the decomposition suffer from incomplete reconstruction, our method preserves the quality of the originally visible regions while achieving most plausible recovery of unknown regions.}
    \label{fig:exp:compare_gen_replica}
    \vspace{1.0 em}
\end{figure*}

\begin{table*}[!t]
\centering
\huge
\setlength{\tabcolsep}{10pt}
\renewcommand{\arraystretch}{0.9}  
\resizebox{0.98 \linewidth}{!}{
\begin{tabular}{ll cccccc ccc}
\specialrule{.1em}{.1em}{.1em}
\multicolumn{2}{c}{\multirow{2}{*}{Methods}} & \multicolumn{2}{c}{PSNR$\uparrow$} & \multicolumn{2}{c}{SSIM$\uparrow$} & \multicolumn{2}{c}{LPIPS$\downarrow$} & \multirow{2}{*}{CD$\downarrow$} & \multirow{2}{*}{F1-Score$\uparrow$} & \multirow{2}{*}{Volume IoU$\uparrow$} \\
\cmidrule[0.5pt](rl){3-4} \cmidrule[0.5pt](rl){5-6} \cmidrule[0.5pt](rl){7-8}  
\multicolumn{2}{c}{}                         & Known      & Unknown     & Known      & Unknown     & Known      & Unknown      &                     &                          &                             \\
\specialrule{.1em}{.1em}{.1em}
Origin Recon.                 & 2DGS~\cite{huang20242d}         & 31.67      & 27.44       & 0.976      & 0.918       & 0.034      & 0.093        & 0.028               & 0.734                    & 0.361                       \\
\midrule
\multirow{2}{*}{Single-View}  & MVDFusion~\cite{hu2024mvd}    & 17.19      & 17.46       & 0.797      & 0.787       & 0.232      & 0.251        & 0.081               & 0.150                    & 0.531                       \\
                              & InstantMesh~\cite{xu2024instantmesh}  & 23.05      & 22.83       & 0.853      & 0.862       & 0.129      & 0.139        & 0.045               & 0.382                    & 0.570                       \\
\midrule
\multirow{2}{*}{Multi-View}   & SparP~\cite{xu2025sparp}        & 25.09      & 23.03       & 0.881      & 0.868       & 0.112      & 0.129        & 0.037               & 0.406                    & 0.590                       \\
                              & Ours         & \textbf{32.57}      & \textbf{29.02}       & \textbf{0.979}      & \textbf{0.944}       & \textbf{0.028}      & \textbf{0.066}        & \textbf{0.016}               & \textbf{0.767}                    & \textbf{0.716}      \\
\specialrule{.1em}{.1em}{.1em}
\end{tabular}
}
\vspace{0.125 em}
\caption{\textbf{Quantitative comparison of In-situ Generation.}
Our in-situ generation achieves superior appearance and geometric alignment with the original scene while ensuring the highest quality recovery in the unknown region.
}
\label{tab:compare_gen}
\vspace{-0.25 em}
\end{table*}

\noindent\textbf{Baselines.}
We compare our approach with the state-of-the-art Gaussian Splatting-based 3D segmentation methods~\cite{qin2024langsplat,ye2025gaussian}. For a fair comparison, 
all methods optimize the feature field based on our pre-trained Gaussian Splatting models and use the same instance segmentation masks~\cite{qi2022high} as 2D semantic priors for supervision.

\noindent\textbf{Comparison Results.}
As shown in Fig.~\ref{fig:exp:lerf_compare}, the mask generated by GSGrouping~\cite{ye2025gaussian} suffers from issues such as floaters and missing segments.
This is mainly due to the inconsistency of its object-tracking,
which is inherently sensitive to mask noise and frame discontinuity. 
LangSplat mislabels repeated instances of similar objects within a scene, such as the bottles in the kitchen in Fig.~\ref{fig:exp:lerf_compare}.  
Both methods exhibit poor feature smoothness and the predicted results contain significant noise, particularly in complex scenarios as shown in Fig.~\ref{fig:exp:zipnerf_compare}. 
In contrast, our method enhances the 2D semantic priors with minimal cost and achieves the most distinguishable feature field, enabling fine-grained instance segmentation even in challenging scenes with frequent adjacent and occluded objects (such as utensils on the shelf in nyc) as demonstrated in Fig.~\ref{fig:exp:zipnerf_compare}.
Please refer to the supplementary material for more results.

\noindent\textbf{Ablation Study.}
To demonstrate the effectiveness of spatial contrastive learning with mutual guidance, we conduct experiments by applying contrastive learning on raw noisy 2D segmentation masks and the preprocessed coarse 3D instance priors.
The qualitative results are presented in Fig.~\ref{fig:motivate}, which shows the improvement of segmentation granularity and feature distinctiveness with mutual guidance.
We also inspect the efficacy of filtered 2D masks and cross-view masks during the spatial contrastive learning in Tab.~\ref{tab:ablation_seg}, which also proves that the design of mutual guidance
significantly improves the distinctiveness of the feature field, leading to more accurate instance-level segmentation.

\vspace{-0.25 em}
\subsection{Instance-Level Complete Reconstruction}
\label{exp:complete}
\vspace{-0.5 em}
In this section, we analyze the instance completion quality of in-situ generation.
As demonstrated in Fig.~\ref{fig:insitu_show}, different from generic reconstruction or image-to-3D tasks, for instance-level complete reconstruction in a certain environment, the reconstructed object should be complete and also align to the real-world scenes with precise geometric shape and close appearance.

\noindent\textbf{Baselines.}
For single-view conditioned generation methods, we compare our approach with MVDFusion~\cite{hu2024mvd} and InstantMesh~\cite{xu2024instantmesh}.
We select the rendering view with the least occlusion and the largest object coverage in our filtered viewpoints as input.
For multi-view conditioned reconstruction methods, we compare our approach with SpaRP~\cite{xu2025sparp}, which implicitly infers the 3D spatial relationships among the given sparse views and uses it to accomplish 3D reconstruction.
For a fair comparison, we use the same optimal views filtered by our method as the multi-view inputs.

\noindent\textbf{Metrics.}
We focus on two aspects, the completion quality and the alignment with the original scene, including both rendering and scale.
Following existing reconstruction~\cite{wang2021neus,huang20242d} and generative methods~\cite{liu2023syncdreamer,hu2024mvd,xu2024instantmesh}, we use Chamfer Distance (CD), F1-Score and Volumetric Intersection over Union (Volume IoU) between GT shapes and instance reconstruction to evaluate the completion quality in unseen regions and assess the spatial alignment.
Additionally, we compute PSNR, SSIM, and LPIPS for both known and unknown views (as set up in Viewpoint Selection) to evaluate the appearance alignment and completion quality.

\noindent\textbf{Datasets.}
We conduct the quantitative comparison on the Replica-CAD Dataset~\cite{szot2021habitat}, which is a synthetic scene composed of artist-recreated scanned objects.
Furthermore, we conduct a qualitative comparison on diverse objects across various real-world complex scenes in the ZipNeRF Dataset~\cite{barron2023zip}.
We also conduct user studies to measure the completion quality, please refer to the supplementary material for details.

\begin{figure*}[!t]
    \centering
    \scriptsize
    % \vspace{-5.5 em}
    \includegraphics[width=\textwidth]{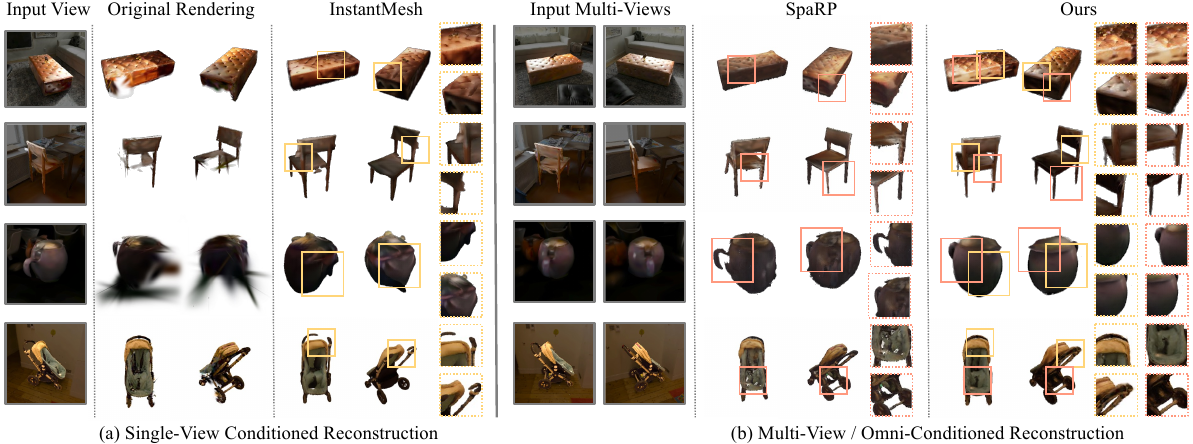}
    \caption{\textbf{Comparison with different generation methods.} 
    We show the complete reconstruction results for each method on diverse instances. Our method not only achieves faithful completion but also maintains consistency with the original scene rendering.
    }
    \label{fig:exp:compare_gen}
    % \vspace{-0.5 em}
\end{figure*}

\begin{figure}[t]
    \centering
    \scriptsize
    \includegraphics[width=0.475 \textwidth]{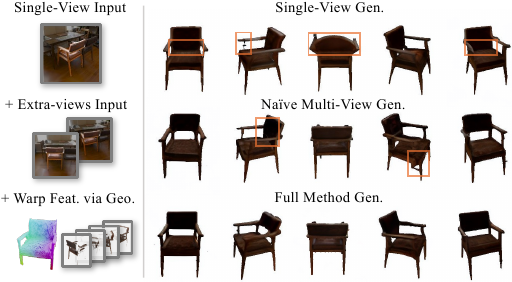}
    \vspace{-1.0 em} 
    \caption{
    \textbf{Ablation studies of the omni-conditioned completion from in-situ generation.}
    Na\"ively alternating the condition views resolves the unrealistic predictions of single-view input, and the geometry-aware feature warping further enhances the consistency of the generated views.
    }
    % \vspace{-0.25 em}
    \label{fig:exp:gen_ab}
\end{figure}

\noindent\textbf{Comparison Results.} 
As shown in Fig.~\ref{fig:exp:compare_gen}, although most methods recover geometry for common objects, such as the leather stool in the first row, only our approach successfully maintains consistent rendering results such as shiny leather texture. 
When reconstructing a partially observed object such as the chair in the second row,
the results from InstantMesh exhibit significant misalignment and contain noticeable floaters. 
In the case of the teapot from the third row, even with multi-view conditions, SpaRP fails to recover the geometry, which is mainly due to the domain gap between the generative priors and real world.
Our method, by employing the warped features from geometric cues,
imposes a geometry-aware constraint and recovers reasonable geometry even under conditions of poor observations.
Moreover, for unusual and complex objects like the baby carriage in the last row, our method recovers the unseen regions while preserving the most realistic rendering quality, whereas other methods suffer from issues such as holes or misalignment.
Fig.~\ref{fig:exp:compare_gen_replica} and metrics shown in Table.~\ref{tab:compare_gen} further demonstrate that our method maintains alignment with the physical world while achieving the most plausible prediction for unseen regions. See more results in supp. material.

\noindent\textbf{Ablation Study.} 
We present the qualitative comparison of our omni-conditioned completion from the in-situ generation in Fig.~\ref{fig:exp:gen_ab}.
The original single-view conditioned method~\cite{hu2024mvd} suffers from inconsistencies between generated views and produces unreasonable results.
The naive multi-view conditioned approach, which alternately
uses additional input views as conditions, avoids unreasonable prediction but still results in floaters and inconsistencies. 
By further adding warped features via geometric cues, which enforces feature consistency from the known observation to the generation, our method significantly enhances the coherence and plausibility of the generated results.

% \vspace{-0.5 em}
\section{Conclusion}
\label{sec:conclusion}
\vspace{-0.25 em}
We have proposed a novel open-set scene decomposition and reconstruction framework, \textbf{\method}.
\method allows users to pick up arbitrary objects from cluttered Gaussian Splatting scenes, and produces complete geometry and appearance of the instance which aligns with the physical world.
The key insight is to trace Gaussian rasterization during mask clustering, which derives spatial contrastive learning that produces highly distinguishable feature fields.
To recover complete objects from occluded observation, we also proposed a novel in-situ generation which fully leverages known information to coordinate the 3D generative prior with faithful instance completion.
As a limitation, we cannot decompose dynamic, transparent or highly reflective objects (see the supplementary material for details).
A potential solution is to incorporate 4D or physically-based representation for these cases, which is left as future work.

\section{Acknowledgments}
\label{sec:acknowledge}
\vspace{-0.5 em}
We express our gratitude to all the anonymous reviewers for their professional and constructive comments. This work was partially supported by the NSFC (No.~62441222), “Innovation Yongjiang 2035” Key R\&D Programme (No.~2025Z062), Information Technology Center, and State Key Lab of CAD\&CG, Zhejiang University.

{
    \small
    \bibliographystyle{ieeenat_fullname}
    \bibliography{main}

\begin{thebibliography}{63}
\providecommand{\natexlab}[1]{#1}
\providecommand{\url}[1]{\texttt{#1}}
\expandafter\ifx\csname urlstyle\endcsname\relax
  \providecommand{\doi}[1]{doi: #1}\else
  \providecommand{\doi}{doi: \begingroup \urlstyle{rm}\Url}\fi

\bibitem[Avetisyan et~al.(2019)Avetisyan, Dahnert, Dai, Savva, Chang, and Nie{\ss}ner]{avetisyan2019scan2cad}
Armen Avetisyan, Manuel Dahnert, Angela Dai, Manolis Savva, Angel~X Chang, and Matthias Nie{\ss}ner.
\newblock Scan2cad: Learning cad model alignment in rgb-d scans.
\newblock In \emph{Proceedings of the IEEE/CVF Conference on computer vision and pattern recognition}, pages 2614--2623, 2019.

\bibitem[Barron et~al.(2023)Barron, Mildenhall, Verbin, Srinivasan, and Hedman]{barron2023zip}
Jonathan~T Barron, Ben Mildenhall, Dor Verbin, Pratul~P Srinivasan, and Peter Hedman.
\newblock Zip-nerf: Anti-aliased grid-based neural radiance fields.
\newblock In \emph{Proceedings of the IEEE/CVF International Conference on Computer Vision}, pages 19697--19705, 2023.

\bibitem[Caron et~al.(2021)Caron, Touvron, Misra, J{\'e}gou, Mairal, Bojanowski, and Joulin]{caron2021emerging}
Mathilde Caron, Hugo Touvron, Ishan Misra, Herv{\'e} J{\'e}gou, Julien Mairal, Piotr Bojanowski, and Armand Joulin.
\newblock Emerging properties in self-supervised vision transformers.
\newblock In \emph{Proceedings of the IEEE/CVF international conference on computer vision}, pages 9650--9660, 2021.

\bibitem[Chang et~al.(2015)Chang, Funkhouser, Guibas, Hanrahan, Huang, Li, Savarese, Savva, Song, Su, et~al.]{chang2015shapenet}
Angel~X Chang, Thomas Funkhouser, Leonidas Guibas, Pat Hanrahan, Qixing Huang, Zimo Li, Silvio Savarese, Manolis Savva, Shuran Song, Hao Su, et~al.
\newblock Shapenet: An information-rich 3d model repository.
\newblock \emph{arXiv preprint arXiv:1512.03012}, 2015.

\bibitem[Choi et~al.(2025)Choi, Song, Kim, Kim, and Do]{choi2025click}
Seokhun Choi, Hyeonseop Song, Jaechul Kim, Taehyeong Kim, and Hoseok Do.
\newblock Click-gaussian: Interactive segmentation to any 3d gaussians.
\newblock In \emph{European Conference on Computer Vision}, pages 289--305. Springer, 2025.

\bibitem[Dong et~al.(2024)Dong, Yang, Ma, Liu, Cui, Bao, Ma, and Cui]{dong2024coin3d}
Wenqi Dong, Bangbang Yang, Lin Ma, Xiao Liu, Liyuan Cui, Hujun Bao, Yuewen Ma, and Zhaopeng Cui.
\newblock Coin3d: Controllable and interactive 3d assets generation with proxy-guided conditioning.
\newblock In \emph{ACM SIGGRAPH 2024 Conference Papers}, pages 1--10, 2024.

\bibitem[Ester et~al.(1996)Ester, Kriegel, Sander, Xu, et~al.]{ester1996density}
Martin Ester, Hans-Peter Kriegel, J{\"o}rg Sander, Xiaowei Xu, et~al.
\newblock A density-based algorithm for discovering clusters in large spatial databases with noise.
\newblock In \emph{kdd}, pages 226--231, 1996.

\bibitem[Fu et~al.(2021)Fu, Cai, Gao, Zhang, Wang, Li, Zeng, Sun, Jia, Zhao, et~al.]{fu20213d}
Huan Fu, Bowen Cai, Lin Gao, Ling-Xiao Zhang, Jiaming Wang, Cao Li, Qixun Zeng, Chengyue Sun, Rongfei Jia, Binqiang Zhao, et~al.
\newblock 3d-front: 3d furnished rooms with layouts and semantics.
\newblock In \emph{Proceedings of the IEEE/CVF International Conference on Computer Vision}, pages 10933--10942, 2021.

\bibitem[Hong et~al.(2023)Hong, Zhang, Gu, Bi, Zhou, Liu, Liu, Sunkavalli, Bui, and Tan]{hong2023lrm}
Yicong Hong, Kai Zhang, Jiuxiang Gu, Sai Bi, Yang Zhou, Difan Liu, Feng Liu, Kalyan Sunkavalli, Trung Bui, and Hao Tan.
\newblock Lrm: Large reconstruction model for single image to 3d.
\newblock \emph{arXiv preprint arXiv:2311.04400}, 2023.

\bibitem[Hu et~al.(2024{\natexlab{a}})Hu, Zhou, Jampani, and Tulsiani]{hu2024mvd}
Hanzhe Hu, Zhizhuo Zhou, Varun Jampani, and Shubham Tulsiani.
\newblock Mvd-fusion: Single-view 3d via depth-consistent multi-view generation.
\newblock In \emph{Proceedings of the IEEE/CVF Conference on Computer Vision and Pattern Recognition}, pages 9698--9707, 2024{\natexlab{a}}.

\bibitem[Hu et~al.(2024{\natexlab{b}})Hu, Ye, Zhao, Lin, He, Wen, He, and Liu]{hu20242}
Yubin Hu, Sheng Ye, Wang Zhao, Matthieu Lin, Yuze He, Yu-Hui Wen, Ying He, and Yong-Jin Liu.
\newblock O\^{} 2-recon: Completing 3d reconstruction of occluded objects in the scene with a pre-trained 2d diffusion model.
\newblock In \emph{Proceedings of the AAAI Conference on Artificial Intelligence}, pages 2285--2293, 2024{\natexlab{b}}.

\bibitem[Huang et~al.(2024)Huang, Yu, Chen, Geiger, and Gao]{huang20242d}
Binbin Huang, Zehao Yu, Anpei Chen, Andreas Geiger, and Shenghua Gao.
\newblock 2d gaussian splatting for geometrically accurate radiance fields.
\newblock In \emph{ACM SIGGRAPH 2024 Conference Papers}, pages 1--11, 2024.

\bibitem[Ji et~al.(2024)Ji, Qiu, Zou, and Wang]{ji2024graspsplats}
Mazeyu Ji, Ri-Zhao Qiu, Xueyan Zou, and Xiaolong Wang.
\newblock Graspsplats: Efficient manipulation with 3d feature splatting.
\newblock \emph{arXiv preprint arXiv:2409.02084}, 2024.

\bibitem[Kerbl et~al.(2023)Kerbl, Kopanas, Leimk{\"u}hler, and Drettakis]{kerbl20233d}
Bernhard Kerbl, Georgios Kopanas, Thomas Leimk{\"u}hler, and George Drettakis.
\newblock 3d gaussian splatting for real-time radiance field rendering.
\newblock \emph{ACM Trans. Graph.}, 42\penalty0 (4):\penalty0 139--1, 2023.

\bibitem[Kerr et~al.(2023)Kerr, Kim, Goldberg, Kanazawa, and Tancik]{kerr2023lerf}
Justin Kerr, Chung~Min Kim, Ken Goldberg, Angjoo Kanazawa, and Matthew Tancik.
\newblock Lerf: Language embedded radiance fields.
\newblock In \emph{Proceedings of the IEEE/CVF International Conference on Computer Vision}, pages 19729--19739, 2023.

\bibitem[Kim et~al.(2024)Kim, Wu, Kerr, Goldberg, Tancik, and Kanazawa]{kim2024garfield}
Chung~Min Kim, Mingxuan Wu, Justin Kerr, Ken Goldberg, Matthew Tancik, and Angjoo Kanazawa.
\newblock Garfield: Group anything with radiance fields.
\newblock In \emph{Proceedings of the IEEE/CVF Conference on Computer Vision and Pattern Recognition}, pages 21530--21539, 2024.

\bibitem[Kirillov et~al.(2023)Kirillov, Mintun, Ravi, Mao, Rolland, Gustafson, Xiao, Whitehead, Berg, Lo, et~al.]{kirillov2023segment}
Alexander Kirillov, Eric Mintun, Nikhila Ravi, Hanzi Mao, Chloe Rolland, Laura Gustafson, Tete Xiao, Spencer Whitehead, Alexander~C Berg, Wan-Yen Lo, et~al.
\newblock Segment anything.
\newblock In \emph{Proceedings of the IEEE/CVF International Conference on Computer Vision}, pages 4015--4026, 2023.

\bibitem[Kroemer et~al.(2021)Kroemer, Niekum, and Konidaris]{kroemer2021review}
Oliver Kroemer, Scott Niekum, and George Konidaris.
\newblock A review of robot learning for manipulation: Challenges, representations, and algorithms.
\newblock \emph{Journal of machine learning research}, 22\penalty0 (30):\penalty0 1--82, 2021.

\bibitem[Li et~al.(2023)Li, Lyu, Ding, Wang, Liao, and Liu]{li2023rico}
Zizhang Li, Xiaoyang Lyu, Yuanyuan Ding, Mengmeng Wang, Yiyi Liao, and Yong Liu.
\newblock Rico: Regularizing the unobservable for indoor compositional reconstruction.
\newblock In \emph{Proceedings of the IEEE/CVF International Conference on Computer Vision}, pages 17761--17771, 2023.

\bibitem[Liu et~al.(2024{\natexlab{a}})Liu, Lin, Liu, Long, Dou, Guo, Luo, and Wang]{liu2024part123}
Anran Liu, Cheng Lin, Yuan Liu, Xiaoxiao Long, Zhiyang Dou, Hao-Xiang Guo, Ping Luo, and Wenping Wang.
\newblock Part123: part-aware 3d reconstruction from a single-view image.
\newblock In \emph{ACM SIGGRAPH 2024 Conference Papers}, pages 1--12, 2024{\natexlab{a}}.

\bibitem[Liu et~al.(2024{\natexlab{b}})Liu, Ye, Nie, He, and Han]{liu2024lasa}
Haolin Liu, Chongjie Ye, Yinyu Nie, Yingfan He, and Xiaoguang Han.
\newblock Lasa: Instance reconstruction from real scans using a large-scale aligned shape annotation dataset.
\newblock In \emph{Proceedings of the IEEE/CVF Conference on Computer Vision and Pattern Recognition}, pages 20454--20464, 2024{\natexlab{b}}.

\bibitem[Liu et~al.(2023{\natexlab{a}})Liu, Zhan, Zhang, Xu, Yu, El~Saddik, Theobalt, Xing, and Lu]{liu2023weakly}
Kunhao Liu, Fangneng Zhan, Jiahui Zhang, Muyu Xu, Yingchen Yu, Abdulmotaleb El~Saddik, Christian Theobalt, Eric Xing, and Shijian Lu.
\newblock Weakly supervised 3d open-vocabulary segmentation.
\newblock \emph{Advances in Neural Information Processing Systems}, 36:\penalty0 53433--53456, 2023{\natexlab{a}}.

\bibitem[Liu et~al.(2023{\natexlab{b}})Liu, Wu, Van~Hoorick, Tokmakov, Zakharov, and Vondrick]{liu2023zero}
Ruoshi Liu, Rundi Wu, Basile Van~Hoorick, Pavel Tokmakov, Sergey Zakharov, and Carl Vondrick.
\newblock Zero-1-to-3: Zero-shot one image to 3d object.
\newblock In \emph{Proceedings of the IEEE/CVF international conference on computer vision}, pages 9298--9309, 2023{\natexlab{b}}.

\bibitem[Liu et~al.(2023{\natexlab{c}})Liu, Lin, Zeng, Long, Liu, Komura, and Wang]{liu2023syncdreamer}
Yuan Liu, Cheng Lin, Zijiao Zeng, Xiaoxiao Long, Lingjie Liu, Taku Komura, and Wenping Wang.
\newblock Syncdreamer: Generating multiview-consistent images from a single-view image.
\newblock \emph{arXiv preprint arXiv:2309.03453}, 2023{\natexlab{c}}.

\bibitem[Liu et~al.(2024{\natexlab{c}})Liu, Ouyang, Wang, Cheng, Xiao, Zhu, Xue, Liu, Shen, and Cao]{liu2024infusion}
Zhiheng Liu, Hao Ouyang, Qiuyu Wang, Ka~Leong Cheng, Jie Xiao, Kai Zhu, Nan Xue, Yu Liu, Yujun Shen, and Yang Cao.
\newblock Infusion: Inpainting 3d gaussians via learning depth completion from diffusion prior.
\newblock \emph{arXiv preprint arXiv:2404.11613}, 2024{\natexlab{c}}.

\bibitem[Lu et~al.(2023)Lu, Chang, Jing, Boularias, and Bekris]{lu2023ovir}
Shiyang Lu, Haonan Chang, Eric~Pu Jing, Abdeslam Boularias, and Kostas Bekris.
\newblock Ovir-3d: Open-vocabulary 3d instance retrieval without training on 3d data.
\newblock In \emph{Conference on Robot Learning}, pages 1610--1620. PMLR, 2023.

\bibitem[Lugmayr et~al.(2022)Lugmayr, Danelljan, Romero, Yu, Timofte, and Van~Gool]{lugmayr2022repaint}
Andreas Lugmayr, Martin Danelljan, Andres Romero, Fisher Yu, Radu Timofte, and Luc Van~Gool.
\newblock Repaint: Inpainting using denoising diffusion probabilistic models.
\newblock In \emph{Proceedings of the IEEE/CVF conference on computer vision and pattern recognition}, pages 11461--11471, 2022.

\bibitem[Mildenhall et~al.(2021)Mildenhall, Srinivasan, Tancik, Barron, Ramamoorthi, and Ng]{mildenhall2021nerf}
Ben Mildenhall, Pratul~P Srinivasan, Matthew Tancik, Jonathan~T Barron, Ravi Ramamoorthi, and Ren Ng.
\newblock Nerf: Representing scenes as neural radiance fields for view synthesis.
\newblock \emph{Communications of the ACM}, 65\penalty0 (1):\penalty0 99--106, 2021.

\bibitem[Mirzaei et~al.(2023)Mirzaei, Aumentado-Armstrong, Derpanis, Kelly, Brubaker, Gilitschenski, and Levinshtein]{mirzaei2023spin}
Ashkan Mirzaei, Tristan Aumentado-Armstrong, Konstantinos~G Derpanis, Jonathan Kelly, Marcus~A Brubaker, Igor Gilitschenski, and Alex Levinshtein.
\newblock Spin-nerf: Multiview segmentation and perceptual inpainting with neural radiance fields.
\newblock In \emph{Proceedings of the IEEE/CVF Conference on Computer Vision and Pattern Recognition}, pages 20669--20679, 2023.

\bibitem[Nguyen et~al.(2024)Nguyen, Ngo, Kalogerakis, Gan, Tran, Pham, and Nguyen]{nguyen2024open3dis}
Phuc Nguyen, Tuan~Duc Ngo, Evangelos Kalogerakis, Chuang Gan, Anh Tran, Cuong Pham, and Khoi Nguyen.
\newblock Open3dis: Open-vocabulary 3d instance segmentation with 2d mask guidance.
\newblock In \emph{Proceedings of the IEEE/CVF Conference on Computer Vision and Pattern Recognition}, pages 4018--4028, 2024.

\bibitem[Ni et~al.(2025)Ni, Liu, Lu, Zhou, Zhu, Chen, and Huang]{ni2025dprecon}
Junfeng Ni, Yu Liu, Ruijie Lu, Zirui Zhou, Song-Chun Zhu, Yixin Chen, and Siyuan Huang.
\newblock Decompositional neural scene reconstruction with generative diffusion prior.
\newblock In \emph{CVPR}, 2025.

\bibitem[Oquab et~al.(2023)Oquab, Darcet, Moutakanni, Vo, Szafraniec, Khalidov, Fernandez, Haziza, Massa, El-Nouby, et~al.]{oquab2023dinov2}
Maxime Oquab, Timoth{\'e}e Darcet, Th{\'e}o Moutakanni, Huy Vo, Marc Szafraniec, Vasil Khalidov, Pierre Fernandez, Daniel Haziza, Francisco Massa, Alaaeldin El-Nouby, et~al.
\newblock Dinov2: Learning robust visual features without supervision.
\newblock \emph{arXiv preprint arXiv:2304.07193}, 2023.

\bibitem[Peng et~al.(2023)Peng, Genova, Jiang, Tagliasacchi, Pollefeys, Funkhouser, et~al.]{peng2023openscene}
Songyou Peng, Kyle Genova, Chiyu Jiang, Andrea Tagliasacchi, Marc Pollefeys, Thomas Funkhouser, et~al.
\newblock Openscene: 3d scene understanding with open vocabularies.
\newblock In \emph{Proceedings of the IEEE/CVF conference on computer vision and pattern recognition}, pages 815--824, 2023.

\bibitem[Poole et~al.(2022)Poole, Jain, Barron, and Mildenhall]{poole2022dreamfusion}
Ben Poole, Ajay Jain, Jonathan~T Barron, and Ben Mildenhall.
\newblock Dreamfusion: Text-to-3d using 2d diffusion.
\newblock \emph{arXiv preprint arXiv:2209.14988}, 2022.

\bibitem[Qi et~al.(2022)Qi, Kuen, Guo, Shen, Gu, Jia, Lin, and Yang]{qi2022high}
Lu Qi, Jason Kuen, Weidong Guo, Tiancheng Shen, Jiuxiang Gu, Jiaya Jia, Zhe Lin, and Ming-Hsuan Yang.
\newblock High-quality entity segmentation.
\newblock \emph{arXiv preprint arXiv:2211.05776}, 2022.

\bibitem[Qin et~al.(2024)Qin, Li, Zhou, Wang, and Pfister]{qin2024langsplat}
Minghan Qin, Wanhua Li, Jiawei Zhou, Haoqian Wang, and Hanspeter Pfister.
\newblock Langsplat: 3d language gaussian splatting.
\newblock In \emph{Proceedings of the IEEE/CVF Conference on Computer Vision and Pattern Recognition}, pages 20051--20060, 2024.

\bibitem[Qiu et~al.(2024)Qiu, Yang, Zeng, and Wang]{qiu2024feature}
Ri-Zhao Qiu, Ge Yang, Weijia Zeng, and Xiaolong Wang.
\newblock Feature splatting: Language-driven physics-based scene synthesis and editing.
\newblock \emph{arXiv preprint arXiv:2404.01223}, 2024.

\bibitem[Radford et~al.(2021)Radford, Kim, Hallacy, Ramesh, Goh, Agarwal, Sastry, Askell, Mishkin, Clark, et~al.]{radford2021learning}
Alec Radford, Jong~Wook Kim, Chris Hallacy, Aditya Ramesh, Gabriel Goh, Sandhini Agarwal, Girish Sastry, Amanda Askell, Pamela Mishkin, Jack Clark, et~al.
\newblock Learning transferable visual models from natural language supervision.
\newblock In \emph{International conference on machine learning}, pages 8748--8763. PMLR, 2021.

\bibitem[Schonberger and Frahm(2016)]{schonberger2016structure}
Johannes~L Schonberger and Jan-Michael Frahm.
\newblock Structure-from-motion revisited.
\newblock In \emph{Proceedings of the IEEE conference on computer vision and pattern recognition}, pages 4104--4113, 2016.

\bibitem[Sch\"{o}nberger et~al.(2016)Sch\"{o}nberger, Zheng, Pollefeys, and Frahm]{schoenberger2016mvs}
Johannes~Lutz Sch\"{o}nberger, Enliang Zheng, Marc Pollefeys, and Jan-Michael Frahm.
\newblock Pixelwise view selection for unstructured multi-view stereo.
\newblock In \emph{European Conference on Computer Vision (ECCV)}, 2016.

\bibitem[Shi et~al.(2024)Shi, Wang, Duan, and Guan]{shi2024language}
Jin-Chuan Shi, Miao Wang, Hao-Bin Duan, and Shao-Hua Guan.
\newblock Language embedded 3d gaussians for open-vocabulary scene understanding.
\newblock In \emph{Proceedings of the IEEE/CVF Conference on Computer Vision and Pattern Recognition}, pages 5333--5343, 2024.

\bibitem[Shi et~al.(2023)Shi, Chen, Zhang, Liu, Xu, Wei, Chen, Zeng, and Su]{shi2023zero123++}
Ruoxi Shi, Hansheng Chen, Zhuoyang Zhang, Minghua Liu, Chao Xu, Xinyue Wei, Linghao Chen, Chong Zeng, and Hao Su.
\newblock Zero123++: a single image to consistent multi-view diffusion base model.
\newblock \emph{arXiv preprint arXiv:2310.15110}, 2023.

\bibitem[Shorinwa et~al.(2024)Shorinwa, Tucker, Smith, Swann, Chen, Firoozi, Kennedy, and Schwager]{shorinwa2024splat}
Olaolu Shorinwa, Johnathan Tucker, Aliyah Smith, Aiden Swann, Timothy Chen, Roya Firoozi, Monroe~David Kennedy, and Mac Schwager.
\newblock Splat-mover: multi-stage, open-vocabulary robotic manipulation via editable gaussian splatting.
\newblock In \emph{8th Annual Conference on Robot Learning}, 2024.

\bibitem[Suvorov et~al.(2022)Suvorov, Logacheva, Mashikhin, Remizova, Ashukha, Silvestrov, Kong, Goka, Park, and Lempitsky]{suvorov2022resolution}
Roman Suvorov, Elizaveta Logacheva, Anton Mashikhin, Anastasia Remizova, Arsenii Ashukha, Aleksei Silvestrov, Naejin Kong, Harshith Goka, Kiwoong Park, and Victor Lempitsky.
\newblock Resolution-robust large mask inpainting with fourier convolutions.
\newblock In \emph{Proceedings of the IEEE/CVF winter conference on applications of computer vision}, pages 2149--2159, 2022.

\bibitem[Szot et~al.(2021)Szot, Clegg, Undersander, Wijmans, Zhao, Turner, Maestre, Mukadam, Chaplot, Maksymets, Gokaslan, Vondrus, Dharur, Meier, Galuba, Chang, Kira, Koltun, Malik, Savva, and Batra]{szot2021habitat}
Andrew Szot, Alex Clegg, Eric Undersander, Erik Wijmans, Yili Zhao, John Turner, Noah Maestre, Mustafa Mukadam, Devendra Chaplot, Oleksandr Maksymets, Aaron Gokaslan, Vladimir Vondrus, Sameer Dharur, Franziska Meier, Wojciech Galuba, Angel Chang, Zsolt Kira, Vladlen Koltun, Jitendra Malik, Manolis Savva, and Dhruv Batra.
\newblock Habitat 2.0: Training home assistants to rearrange their habitat.
\newblock In \emph{Advances in Neural Information Processing Systems (NeurIPS)}, 2021.

\bibitem[Takmaz et~al.(2023)Takmaz, Fedele, Sumner, Pollefeys, Tombari, and Engelmann]{takmaz2023openmask3d}
Ay{\c{c}}a Takmaz, Elisabetta Fedele, Robert~W Sumner, Marc Pollefeys, Federico Tombari, and Francis Engelmann.
\newblock Openmask3d: Open-vocabulary 3d instance segmentation.
\newblock \emph{arXiv preprint arXiv:2306.13631}, 2023.

\bibitem[Wang et~al.(2021)Wang, Liu, Liu, Theobalt, Komura, and Wang]{wang2021neus}
Peng Wang, Lingjie Liu, Yuan Liu, Christian Theobalt, Taku Komura, and Wenping Wang.
\newblock Neus: Learning neural implicit surfaces by volume rendering for multi-view reconstruction.
\newblock \emph{arXiv preprint arXiv:2106.10689}, 2021.

\bibitem[Weber et~al.(2024)Weber, Holynski, Jampani, Saxena, Snavely, Kar, and Kanazawa]{weber2024nerfiller}
Ethan Weber, Aleksander Holynski, Varun Jampani, Saurabh Saxena, Noah Snavely, Abhishek Kar, and Angjoo Kanazawa.
\newblock Nerfiller: Completing scenes via generative 3d inpainting.
\newblock In \emph{Proceedings of the IEEE/CVF Conference on Computer Vision and Pattern Recognition}, pages 20731--20741, 2024.

\bibitem[Wu et~al.(2022)Wu, Liu, Chen, Li, Zheng, Cai, and Zheng]{wu2022object}
Qianyi Wu, Xian Liu, Yuedong Chen, Kejie Li, Chuanxia Zheng, Jianfei Cai, and Jianmin Zheng.
\newblock Object-compositional neural implicit surfaces.
\newblock In \emph{European Conference on Computer Vision}, pages 197--213. Springer, 2022.

\bibitem[Wu et~al.(2023)Wu, Wang, Li, Zheng, and Cai]{wu2023objectsdf++}
Qianyi Wu, Kaisiyuan Wang, Kejie Li, Jianmin Zheng, and Jianfei Cai.
\newblock Objectsdf++: Improved object-compositional neural implicit surfaces.
\newblock In \emph{Proceedings of the IEEE/CVF International Conference on Computer Vision}, pages 21764--21774, 2023.

\bibitem[Wu et~al.(2024)Wu, Meng, Li, Wu, Shi, Cheng, Zhao, Feng, Ding, Wang, et~al.]{wu2024opengaussian}
Yanmin Wu, Jiarui Meng, Haijie Li, Chenming Wu, Yahao Shi, Xinhua Cheng, Chen Zhao, Haocheng Feng, Errui Ding, Jingdong Wang, et~al.
\newblock Opengaussian: Towards point-level 3d gaussian-based open vocabulary understanding.
\newblock \emph{arXiv preprint arXiv:2406.02058}, 2024.

\bibitem[Xu et~al.(2025)Xu, Li, Chen, Liu, Shi, Su, and Liu]{xu2025sparp}
Chao Xu, Ang Li, Linghao Chen, Yulin Liu, Ruoxi Shi, Hao Su, and Minghua Liu.
\newblock Sparp: Fast 3d object reconstruction and pose estimation from sparse views.
\newblock In \emph{European Conference on Computer Vision}, pages 143--163. Springer, 2025.

\bibitem[Xu et~al.(2024)Xu, Cheng, Gao, Wang, Gao, and Shan]{xu2024instantmesh}
Jiale Xu, Weihao Cheng, Yiming Gao, Xintao Wang, Shenghua Gao, and Ying Shan.
\newblock Instantmesh: Efficient 3d mesh generation from a single image with sparse-view large reconstruction models.
\newblock \emph{arXiv preprint arXiv:2404.07191}, 2024.

\bibitem[Yan et~al.(2024{\natexlab{a}})Yan, Zhang, Zhu, and Wang]{yan2024maskclustering}
Mi Yan, Jiazhao Zhang, Yan Zhu, and He Wang.
\newblock Maskclustering: View consensus based mask graph clustering for open-vocabulary 3d instance segmentation.
\newblock In \emph{Proceedings of the IEEE/CVF Conference on Computer Vision and Pattern Recognition}, pages 28274--28284, 2024{\natexlab{a}}.

\bibitem[Yan et~al.(2024{\natexlab{b}})Yan, Lin, Zhou, Wang, Sun, Zhan, Lang, Zhou, and Peng]{yan2024street}
Yunzhi Yan, Haotong Lin, Chenxu Zhou, Weijie Wang, Haiyang Sun, Kun Zhan, Xianpeng Lang, Xiaowei Zhou, and Sida Peng.
\newblock Street gaussians for modeling dynamic urban scenes.
\newblock \emph{arXiv preprint arXiv:2401.01339}, 2024{\natexlab{b}}.

\bibitem[Yang et~al.(2021)Yang, Zhang, Xu, Li, Zhou, Bao, Zhang, and Cui]{yang2021objectnerf}
Bangbang Yang, Yinda Zhang, Yinghao Xu, Yijin Li, Han Zhou, Hujun Bao, Guofeng Zhang, and Zhaopeng Cui.
\newblock Learning object-compositional neural radiance field for editable scene rendering.
\newblock In \emph{International Conference on Computer Vision ({ICCV})}, 2021.

\bibitem[Yang et~al.(2023)Yang, Wu, He, Zhao, and Liu]{yang2023sam3d}
Yunhan Yang, Xiaoyang Wu, Tong He, Hengshuang Zhao, and Xihui Liu.
\newblock Sam3d: Segment anything in 3d scenes.
\newblock \emph{arXiv preprint arXiv:2306.03908}, 2023.

\bibitem[Ye et~al.(2025)Ye, Danelljan, Yu, and Ke]{ye2025gaussian}
Mingqiao Ye, Martin Danelljan, Fisher Yu, and Lei Ke.
\newblock Gaussian grouping: Segment and edit anything in 3d scenes.
\newblock In \emph{European Conference on Computer Vision}, pages 162--179. Springer, 2025.

\bibitem[Yin et~al.(2024)Yin, Liu, Xiao, Cohen-Or, Huang, and Chen]{yin2024sai3d}
Yingda Yin, Yuzheng Liu, Yang Xiao, Daniel Cohen-Or, Jingwei Huang, and Baoquan Chen.
\newblock Sai3d: Segment any instance in 3d scenes.
\newblock In \emph{Proceedings of the IEEE/CVF Conference on Computer Vision and Pattern Recognition}, pages 3292--3302, 2024.

\bibitem[Ying et~al.(2024)Ying, Yin, Zhang, Wang, Yu, Huang, and Fang]{ying2024omniseg3d}
Haiyang Ying, Yixuan Yin, Jinzhi Zhang, Fan Wang, Tao Yu, Ruqi Huang, and Lu Fang.
\newblock Omniseg3d: Omniversal 3d segmentation via hierarchical contrastive learning.
\newblock In \emph{Proceedings of the IEEE/CVF Conference on Computer Vision and Pattern Recognition}, pages 20612--20622, 2024.

\bibitem[Zhang et~al.(2024)Zhang, Wang, Zhang, Qiu, Pang, Jiang, Yang, Xu, and Yu]{zhang2024clay}
Longwen Zhang, Ziyu Wang, Qixuan Zhang, Qiwei Qiu, Anqi Pang, Haoran Jiang, Wei Yang, Lan Xu, and Jingyi Yu.
\newblock Clay: A controllable large-scale generative model for creating high-quality 3d assets.
\newblock \emph{ACM Transactions on Graphics (TOG)}, 43\penalty0 (4):\penalty0 1--20, 2024.

\bibitem[Zhou et~al.(2024{\natexlab{a}})Zhou, Chang, Jiang, Fan, Zhu, Xu, Chari, You, Wang, and Kadambi]{zhou2024feature}
Shijie Zhou, Haoran Chang, Sicheng Jiang, Zhiwen Fan, Zehao Zhu, Dejia Xu, Pradyumna Chari, Suya You, Zhangyang Wang, and Achuta Kadambi.
\newblock Feature 3dgs: Supercharging 3d gaussian splatting to enable distilled feature fields.
\newblock In \emph{Proceedings of the IEEE/CVF Conference on Computer Vision and Pattern Recognition}, pages 21676--21685, 2024{\natexlab{a}}.

\bibitem[Zhou et~al.(2024{\natexlab{b}})Zhou, Lin, Shan, Wang, Sun, and Yang]{zhou2024drivinggaussian}
Xiaoyu Zhou, Zhiwei Lin, Xiaojun Shan, Yongtao Wang, Deqing Sun, and Ming-Hsuan Yang.
\newblock Drivinggaussian: Composite gaussian splatting for surrounding dynamic autonomous driving scenes.
\newblock In \emph{Proceedings of the IEEE/CVF Conference on Computer Vision and Pattern Recognition}, pages 21634--21643, 2024{\natexlab{b}}.

\end{thebibliography}
}

\end{document}